%% file: main.tex
\documentclass[10pt,twocolumn,letterpaper]{article}
\usepackage{iccv}   

\definecolor{iccvblue}{rgb}{0.21,0.49,0.74}
\usepackage[pagebackref,breaklinks,colorlinks,allcolors=iccvblue]{hyperref}
\usepackage[accsupp]{axessibility}  
\usepackage{graphicx}  
\usepackage{array}     
\usepackage{amssymb}
\usepackage{xcolor}
\usepackage{bbding}
\usepackage{pifont}
\usepackage{colortbl}
\usepackage{bbding}  
\usepackage{multirow}

\def\paperID{13753} 
\def\confName{ICCV}
\def\confYear{2025}

\title{ObjectRelator: Enabling Cross-View Object Relation Understanding \\  Across Ego-Centric and Exo-Centric Perspectives}
\author{\textbf{Yuqian Fu}$^{1}$\thanks{Correspondence to \textit{yuqian.fu@insait.ai}.}\quad \textbf{Runze Wang}$^{2}$\quad \textbf{Bin Ren}$^{1,3,4}$\quad \textbf{Guolei Sun}$^{5}$\quad \textbf{Biao Gong}$^{6}$ \\ \textbf{Yanwei Fu}$^{2}$\quad \textbf{Danda Pani Paudel}$^{1}$\quad \textbf{Xuanjing Huang}$^{2}$\quad \textbf{Luc Van Gool}$^{1}$  \\
\small $^1$INSAIT, Sofia University ``St. Kliment Ohridski", $^2$Fudan University, $^3$University of Trento, $^4$University of Pisa, $^5$ETH Zürich, $^6$Ant Group \\
}

\begin{document}
\maketitle

\begin{abstract}
Bridging the gap between ego-centric and exo-centric views has been a long-standing question in computer vision. In this paper, we focus on the emerging Ego-Exo object correspondence task, which aims to understand object relations across ego-exo perspectives through segmentation. While numerous segmentation models have been proposed, most operate on a single image (view), making them impractical for cross-view scenarios. 
PSALM~\cite{zhang2024psalm}, a recently proposed segmentation method, stands out as a notable exception with its demonstrated zero-shot ability on this task. 
However, due to the drastic viewpoint change between ego and exo, PSALM fails to accurately locate and segment objects, especially in complex backgrounds or when object appearances change significantly.
To address these issues, we propose ObjectRelator, a novel approach featuring two key modules: Multimodal Condition Fusion (MCFuse) and SSL-based Cross-View Object Alignment (XObjAlign). MCFuse introduces language as an additional cue, integrating both visual masks and textual descriptions to improve object localization and prevent incorrect associations. XObjAlign enforces cross-view consistency through self-supervised alignment, enhancing robustness to object appearance variations. 
Extensive experiments demonstrate ObjectRelator’s effectiveness on the large-scale Ego-Exo4D benchmark and HANDAL-X (an adapted dataset for cross-view segmentation) with state-of-the-art performance. 
Code is available at: \href{http://yuqianfu.com/ObjectRelator/}{http://yuqianfu.com/ObjectRelator}. 
\end{abstract}

\section{Introduction}
\label{sec:intro}
Advancements in vision tasks~\cite{karpathy2014large,xie2018rethinking,ouyang2024codef,krishna2017dense, iashin2020multi,yang2023vid2seq} have largely focused on exo-centric (third-person) perspectives. 
In contrast, progress on ego-centric (first-person) perception is relatively behind. 
To bridge this gap, most efforts have focused on developing ego-centric models~\cite{kitani2011fast, damen2018scaling, lin2022egocentric, grauman2022ego4d, zhang2023helping, wang2023holoassist} while fewer studies~\cite{regmi2019cross, tang2019multi, ren2021cascaded, luo2025put} have directly explored the relationships between the two views, primarily due to the lack of cross-view datasets. 
The recent release of the Ego-Exo4D~\cite{grauman2024ego}, a large-scale, temporally aligned ego-exo dataset, enables the study of \textit{cross-view object relation understanding}, offering a promising new direction.

\begin{figure}[t]
    \centering
    {\includegraphics[width=0.95\linewidth]{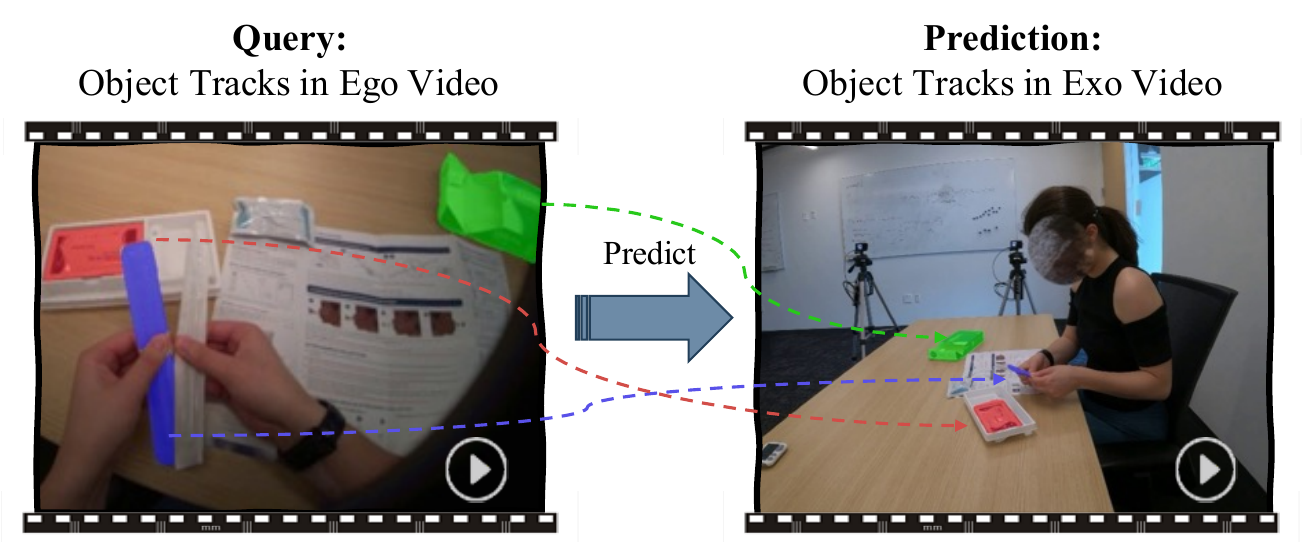}}
    \caption{Illustration of the Ego-Exo Object Correspondence Task (example shown: Ego2Exo).}
    \vspace{-0.18in}
    \label{fig:teaser}
\end{figure}

Specifically, in this paper, we tackle the task of \textbf{Ego-Exo Object Correspondence}\footnote{Originally termed ``Ego-Exo Correspondence" in Ego-Exo4D~\cite{grauman2024ego}. Renamed for clarity, emphasizing object-level mapping.}, first introduced in Ego-Exo4D~\cite{grauman2024ego}. 
As shown in Fig.\ref{fig:teaser}, given object queries from one perspective (\textit{e.g.}, ego view), the task involves predicting the corresponding object masks in another perspective (\textit{e.g.}, exo view). Solving this task unlocks new possibilities in VR and robotics, \textit{e.g.}, enabling virtual agents or robots to manipulate ego-view actions by learning from exo-view demonstrations. 
However, despite the existence of numerous segmentation models, this task remains non-trivial: most prior approaches operate on a single image (view), whereas Ego-Exo Object Correspondence requires leveraging visual cues from one perspective to identify corresponding objects in a completely different one.

Critically, PSALM~\cite{zhang2024psalm}, a recently proposed vision-language-based segmentation method, stands out as a notable exception. 
Combining Mask2Former~\cite{cheng2022masked} with the powerful Large language model (LLM), PSALM outperforms flagship segmentation models such as LISA~\cite{lai2024lisa} and SEEM~\cite{zou2024segment}, while also enabling zero-shot learning (ZSL) for the ego-exo object correspondence task.
However, despite achieving state-of-the-art performance on other segmentation tasks, PSALM struggles with the unique challenges of ego-exo object correspondence: 
i) Complex backgrounds in the exo view often introduce distractors, causing the model to segment objects with similar shapes but incorrect categories; 
ii) Significant variations in object appearance between ego and exo views further increase the task’s difficulty.
These challenges motivate us to enhance PSALM by:
\ding{172} \textit{improving object localization} and 
\ding{173} \textit{ensuring object consistency across perspectives}.

To this end, we propose two innovative modules, \textit{i.e.}, Multimodal Condition Fusion (MCFuse) and Cross-View Object Alignment (XObjAlign).
MCFuse addresses \ding{172} by incorporating language descriptions into the task. 
While the benchmark provides only object query masks, MCFuse managed to generate textual descriptions of the query object via large vision-language models like LLaVA~\cite{liu2023llava}. Then, a lightweight cross-attention with residual connections and learnable fusion weights fuses the visual mask and generated text in the embedding space. 
This enables the model to leverage both visual and linguistic cues, improving target object localization. 
To tackle \ding{173}, XObjAlign is proposed to align paired ego and exo objects in a shared latent space, enforcing proximity between them. 
Employing a self-supervised learning approach, we optimize the similarity between ego and exo representations, particularly at the object level, improving the model's ability to recognize objects across different perspectives and address challenges arising from viewpoint variations.

By integrating the proposed MCFuse and XObjAlign into PSALM, we propose our \textbf{ObjectRelator}. We validate its effectiveness through extensive experiments on the large-scale Ego-Exo4D benchmark, achieving state-of-the-art (SOTA) performance with significant improvements over existing competitors. 
Additionally, we introduce HANDAL-X, an adapted cross-view object segmentation dataset with robotics-ready, manipulable objects as a new testbed, and demonstrate our method's superiority.

We summarize our contributions as:

$\bullet$ \textbf{Toward Ego-Exo Object Correspondence Task:} We conduct an early exploration of this challenging task, analyzing its unique difficulties, constructing several baselines, and proposing a new method.

$\bullet$ \textbf{ObjectRelator Framework:} We introduce ObjectRelator, a cross-view object segmentation method combining MCFuse and XObjAlign. MCFuse for the first time introduces the text modality into this task and improves localization using multimodal cues for the same object(s), while XObjAlign boosts performance under appearance variations with an object-level consistency constraint.

$\bullet$ \textbf{New Testbed $\&$ SOTA Results:} Alongside Ego-Exo4D, we present HANDAL-X as an additional benchmark. Our proposed ObjectRelator achieves state-of-the-art (SOTA) results on both datasets.

\section{Related Work}
\label{sec:related}
\noindent\textbf{Ego-Exo Perception Understanding.} 
Both ego- and exo-centric perceptions are crucial for understanding the world. Extensive work has been done in single-view understanding: For exo-centric view, numerous methods address tasks such as video classification~\cite{karpathy2014large, xie2018rethinking, fu2019embodied, fu2020depth, ouyang2024codef}, captioning~\cite{krishna2017dense, iashin2020multi, yang2023vid2seq}, detection and segmentation~\cite{tan2024xtrack, cheng2023tracking, brodermann2025cafuser, zhou2025camsam2}, and generation~\cite{tulyakov2018mocogan, singer2022make, wu2023tune, huang2024learning}. While ego-centric research has historically lagged behind exo-centric, recent efforts including datasets such as Ego4D~\cite{grauman2022ego4d}, EK100~\cite{damen2018scaling}, HoloAssist~\cite{wang2023holoassist}, AssistQ~\cite{wong2022assistq}, as well as a variety of methods~\cite{li2015delving, kitani2011fast, fathi2011understanding, jia2022egotaskqa, lin2022egocentric, zhang2023helping, pramanick2023egovlpv2, zhang2024object} have brought substantial progress.
However, far fewer studies directly address the ego-exo relationships. Notable exceptions have ego-exo translation/generation~\cite{tang2019multi, luo2025put, cheng20254diff, luo2024intention, liuexocentric}. 
One reason for this gap is that existing datasets are often small or lack comprehensive annotations~\cite{de2009guide, sigurdsson2018charades, jia2020lemma, rai2021home, kwon2021h2o, Huang_2024_CVPR, sener2022assembly101}.
The recent release of Ego-Exo4D~\cite{grauman2024ego}, a large-scale, richly annotated, and time-synchronized dataset, opens new chances for advancing ego-exo research. In this paper, we focus on the ego-exo object correspondence task proposed in Ego-Exo4D, aiming to establish a good starting point.

\begin{table*}[t]
    \centering
    \setlength\tabcolsep{5pt}
    \scalebox{0.67}{
    \begin{tabular}{c|c|c|c|c}
    \toprule[0.95pt]
    \textbf{Methods}           &   \textbf{Task}         & \textbf{Extra Prompts}                                                                                                        & \textbf{Cross-View} & \textbf{Comparable to Ours} \\ \midrule[0.6pt]
    Mask-RCNN~\cite{he2017mask}, Mask2Former~\cite{cheng2022masked}     & Generic Segmentation             &\ding{55}                                                                                                                         &\ding{55}                 &\ding{55}                         \\ \midrule[0.6pt]
    LISA~\cite{lai2024lisa}, GLAMM~\cite{rasheed2024glamm}, PixelLM~\cite{ren2024pixellm}  & {Referring Segmentation}          & Text Descriptions                                                                                                             &\ding{55}                 &\ding{55}                         \\ \midrule[0.6pt]
    SAM~\cite{kirillov2023segment}, OMG-Seg~\cite{li2024omg}, SAM2~\cite{ravi2024sam2}    & {Interactive Segmentation}             & Visual Prompts,  Prompt Decoder w/o referring to image                                                                     &\ding{55}                 &\ding{55}                         \\ \midrule[0.6pt]
    UNINEXT~\cite{yan2023universal}, SEEM~\cite{zou2024segment}, PSALM~\cite{zhang2024psalm}   & {Universal Segmentation}        & Text or Visual Prompts,  Prompt Decoder with referring to image                                                                           &\ding{55}                 &\ding{52}                         \\ 
    \midrule[0.6pt]
    \cellcolor{cyan!10}\textbf{ObjectRelator (Ours)}   & \cellcolor{cyan!10}\textbf{Ego-Exo Object Correspondence} &  \cellcolor{cyan!10}Visual Prompts,  Prompt Decoder with referring to image  &\cellcolor{cyan!10}\ding{52}                & \cellcolor{cyan!10}-   \\
    \bottomrule[0.95pt]
    \end{tabular}
    }
    \caption{Comparison of tasks and methods. Note that prompt decoder choice depends on the method, not necessarily linked to the task.}
    \vspace{-0.1in}
    \label{tab:relatedwork}
\end{table*}

\noindent\textbf{Segmentation Models.} 
Segmentation models can roughly be grouped into several categories based on input types: we refer to \textit{generic segmentation}~\cite{minaee2021image} as those more classical segmentation tasks e.g., semantic segmentation~\cite{lateef2019survey}, instance segmentation~\cite{hafiz2020survey}, and panoptic segmentation~\cite{kirillov2019panoptic}. Generic segmentation methods (e.g., Mask R-CNN~\cite{he2017mask} and Mask2Former~\cite{cheng2022masked}) don't need extra text prompts or visual prompts. In contrast, \textit{referring segmentation}~\cite{ji2024survey} requires textual descriptions to guide the segmentation. Flagship examples include LISA~\cite{lai2024lisa}, GLAMM~\cite{rasheed2024glamm}, PixelLM~\cite{ren2024pixellm}. \textit{Interactive segmentation}~\cite{ramadan2020survey} utilizes visual hints such as bounding boxes, key points, or masks provided within the image as prompts. Well-known examples have SAM~\cite{kirillov2023segment}, OMG-Seg~\cite{li2024omg}, SAM2~\cite{ravi2024sam2}. The latest advancements such as UNINEXT~\cite{yan2023universal}, SEEM~\cite{zou2024segment}, PSALM~\cite{zhang2024psalm} accept multiple conditions as input, which we termed as \textit{universal segmentation}. However, none of them are specifically tailored for the ego-exo object correspondence, which is essentially challenged by the huge view change between two images.

More differences of methods are summarized in Tab.\ref{tab:relatedwork}. Typically, SAM~\cite{kirillov2023segment}, OMG-Seg~\cite{li2024omg}, and SAM2~\cite{ravi2024sam2} decode visual prompts solely from hints such as points, masks, and bounding boxes, without referencing the corresponding image. These approaches require strict alignment between the mask and testing image, limiting adaptability for tasks like ours, where object locations and shapes vary inherently. By contrast, UNINEXT~\cite{yan2023universal}, SEEM~\cite{zou2024segment}, and PSALM~\cite{zhang2024psalm} decode visual prompts with image reference, making them more comparable to us. Importantly, we differ from existing segmentation approaches by focusing on cross-view challenges and exploring the fusion of multiple conditions.

\noindent\textbf{Cross-View Alignment.}
Tackling data with related but different views has been an ongoing challenge, explored across various domains. For example, person re-identification methods~\cite{wang2015zero, lin2018multi, jia2021unsupervised, zhang2024multi} aim to improve consistency for the same person across different cameras, cross-view image matching/retrieval~\cite{shi2022beyond, tian2017cross, hu2018cvm} seeks to align images captured from different angles, and cross-view image translation~\cite{isola2017image, tang2019multi, li2024crossviewdiff} requires models to generate a novel view of an image while preserving its content. 
Image co-segmentation, such as SegSwap~\cite{shen2021learning}, which segments shared objects across various images, and video object tracking, such as XMem~\cite{cheng2022xmem}, which tracks objects across frames, seem to be the most closely related tasks to us. However, while these tasks share some similarities, we address fundamentally different challenges and employ distinct methodologies in our work.

\section{Methodology}
\label{sec:method}
\noindent\textbf{Task Formulation}.
The \textit{Ego-Exo Object Correspondence} involves mapping objects between ego-exo perspectives as in Fig.~\ref{fig:teaser}. Formally, given a query view $I^{*}/V^{*}$ (e.g., an ego-centric or exo-centric image/video) and a target view $I/V$ (e.g., an exo-centric or ego-centric image/video), along with object mask(s) $m^{*}$ indicating the object of interest in query view, the objective is to predict the corresponding object mask(s) in the target view $m$. The query and target images/videos are assumed to be temporally aligned, ensuring that corresponding frames represent the same time points. Additionally, each pair may contain multiple objects of interest. Any semantic information (e.g., the categories of the objects) is not originally provided in the benchmark. 

To structure and simplify usage, we rename the two sub-tasks as \textbf{Ego2Exo}, where the ego-centric view serves as the query and the exo-centric view as the target, and \textbf{Exo2Ego}, where the roles of ego and exo are reversed.

\noindent\textbf{PSALM Baseline}. 
As discussed in Sec.~\ref{sec:related} and shown in Tab.~\ref{tab:relatedwork}, the UNINEXT~\cite{yan2023universal}, SEEM~\cite{zou2024segment}, PSALM~\cite{zhang2024psalm} could be considered as the baselines. Among them,  we select PSALM for its: 1) validated superior performance over UNINEXT and SEEM models on several various benchmarks; 2) demonstrated zero-shot learning (ZSL) transfer ability to our task. Particularly,  PSALM combines Mask2Former~\cite{cheng2022masked}, the potent generic segmentation model, with a powerful LLM leveraging the strengths of both. 

Briefly, PSALM mainly contains a Visual Encoder, MM (Multimodal) Projector, LLM, Pixel Decoder, Mask Generator. For each segmentation, PSALM takes an image, an instruction prompt, a condition prompt (e.g., categories, text, or visual hints), and mask tokens as input, feeds their corresponding feature or tokens into the LLM resulting in the condition embedding, mask embedding. After that,  PSALM predicts masks in a Mask2Former style: it inputs the multiscale image feature, condition embedding, and mask embedding to the Mask Generator. The image feature for LLM is extracted by the Visual Encoder and MM Projector, while the multiscale image feature for the Mask Generator is extracted by the  Visual Encoder and Pixel Decoder. During the training, loss $\mathcal{L}_{mask}$ from Mask Generator is used to optimize all the modules except the visual encoder; During inference, the Mask Generator generates the predicted mask. For more details, e.g., token extraction method and loss function, please refer to PSALM~\cite{zhang2024psalm}.

\begin{figure*}[t]
    \centering
       {\includegraphics[width=0.9\linewidth]{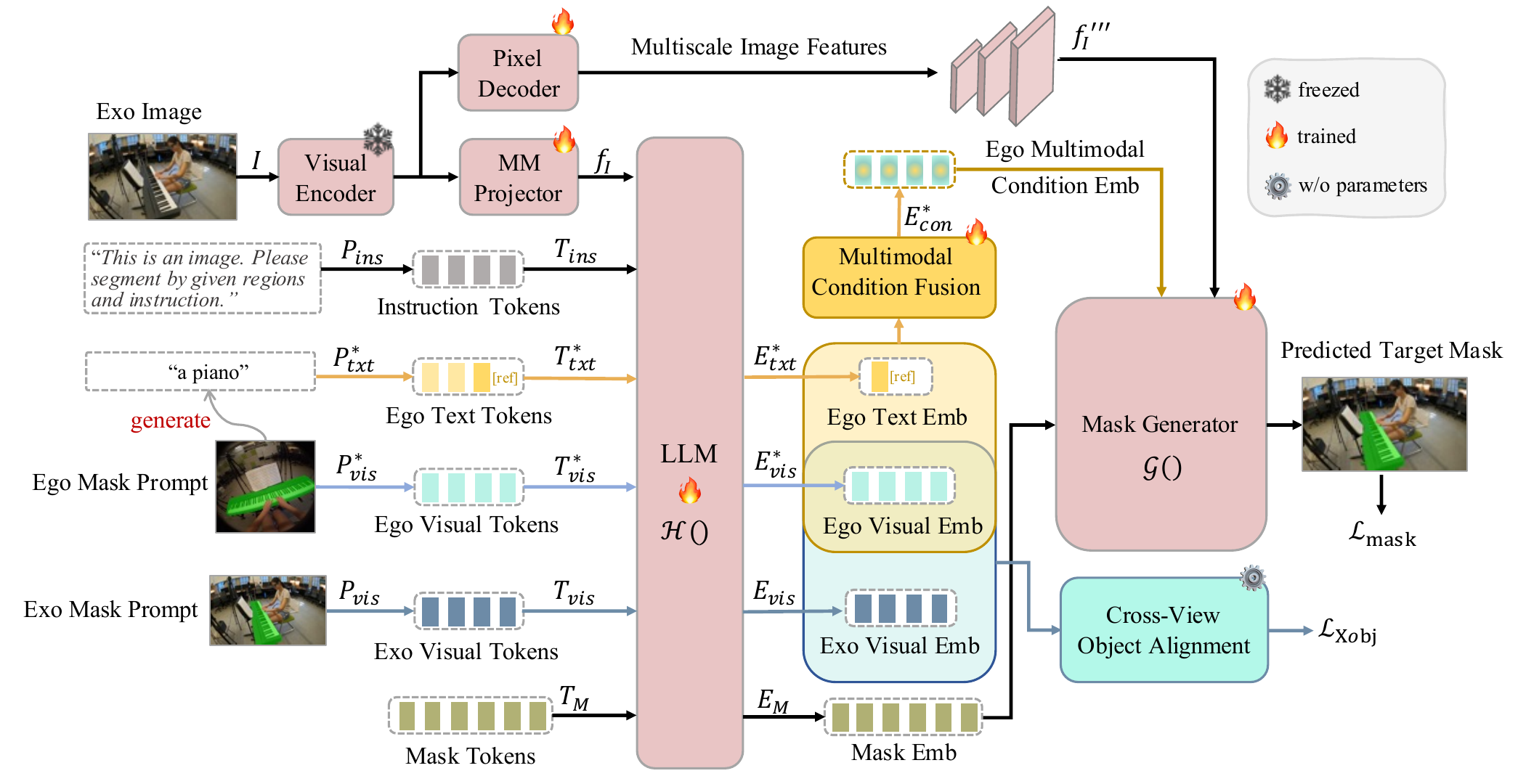}}
    \vspace{-0.05in}
    \caption{\textbf{Overview of ObjectRelator.} Ego2Exo is used as an example. Our method builds on the PSALM baseline (pink blocks) and tailors it for Ego-Exo Object Correspondence with two novel modules: Multimodal Condition Fusion and Cross-View Object Alignment.}
    \vspace{-0.1in}
    \label{fig:framework}
\end{figure*}

\subsection{ObjectRelator Framework} \label{sec:framework}
Our ObjectRelator operates at the frame level, focusing on studying object relations between ego-exo frames, leaving the exploration of temporal information as future work. The framework is illustrated in Fig.~\ref{fig:framework}. 
Overall, we retain the core PSALM modules: Visual Encoder, MM (Multimodal) Projector, LLM $\mathcal{H}()$, Pixel Decoder, Mask Generator $\mathcal{G}()$, loss function $\mathcal{L}_{mask}$, and also its way of extract tokens, while introducing two novel components, \textit{Multimodal Condition Fusion (MCFuse)} and \textit{Cross-View Object Alignment (XObjAlign)}, highlighted in orange and light green colors.

For clarity, we use Ego2Exo as an example: each training step involves a synchronized ego image as the query and an exo image as the target. For each object in the paired data, we denote the ego image, ego object mask, exo image, and exo object mask as $I^{*}$, $m^{*}$, $I$, and $m$ (ground truth), respectively. The following steps are:

\noindent \textbf{1) Preparing Inputs for LLM:} 
For the target exo image $I$, we use the Visual Encoder, MM Projector, and Pixel Decoder to extract features $f_{I}$ and $f_{I}^{'''}$, where $f_{I}^{'''}$ indicates the feature has multiple layers. Unlike the ZSL PSALM model, which only uses the single ego object mask as a prompt for exo, we leverage both the visual object region and a language description as prompts. Thus, our instruction prompt $P_{ins}$ is formulated as: ``\textit{This is an image. Please segment by given regions and instruction.}", where the region hint is ego mask prompt $P_{vis}^{*}$ and the instruction hint is the text instruction prompt $P_{txt}^{*}$. Specifically, $P_{vis}^{*}$ is formed by combing the ego image $I^{*}$ and its object mask $m^{*}$, while $P_{txt}^{*}$ describes the masked object in $I^{*}$, generated via the pre-trained multimodal model LLaVA~\cite{liu2023llava}. Additionally, to enforce consistency between ego and exo objects, we use the ground truth $m$ to form the exo mask prompt $P_{vis}$. Note that although named a prompt, $P_{vis}$ is not directly used for mask generation but mainly for the XObjAlign module during training. We then extract the instruction tokens $T_{ins}$, ego text condition token $T_{txt}^{*}$, ego visual condition token $T_{vis}^{*}$, exo visual token $T_{vis}$, and mask token $T_{M}$, following the same token extraction method as the base PSALM. 

\noindent \textbf{2) Forwarding the Network:} 
We process image features $f_{I}$, $f_{I}^{'''}$, instruction tokens $T_{ins}$, ego text tokens $T_{txt}^{*}$, ego visual tokens $T_{vis}^{*}$, exo visual tokens $T_{vis}$, and mask tokens $T_{M}$. To feed them into the pretrained LLM $\mathcal{H}()$, we create two inputs: one concatenating $f_{I}$, $T_{ins}$, $T_{txt}^{*}$, $T_{vis}^{*}$, and $T_{M}$, and the other with $f_{I}$, $T_{ins}$, $T_{txt}^{*}$, $T_{vis}$, and $T_{M}$. Input both to the LLM yields ego text emb $E_{txt}^{*}$, ego visual emb $E_{vis}^{*}$, exo visual emb $E_{vis}$, and mask emb $E_M$ as,
\begin{gather}
E_{txt}^{*}, E_{vis}^{*}, E_M = \mathcal{H}(f_{I}, T_{ins}, T_{txt}^{*}, T_{vis}^{*}, T_{M}). \\
\rule{1.5em}{0.5pt}, E_{vis}, \rule{1.5em}{0.5pt} = \mathcal{H}(f_{I}, T_{ins}, T_{txt}^{*}, T_{vis}, T_{M}).
\end{gather}

The embeddings $E_{vis}^{*}$ and $E_{vis}$, representing the same object from different views, are passed through the XObjAlign module to compute the cross-object consistency loss $\mathcal{L}_{X{obj}}$. Meanwhile, $E_{txt}^{*}$ and $E_{vis}^{*}$ are sent to the MCFuse module to generate the fused ego multimodal condition embedding $E_{con}^{*}$. Finally, the multiscale image feature $f_{I}^{'''}$, the fused ego multimodal emb $E_{con}^{*}$, and the mask emb $E_M$ are passed to the Mask Generator $\mathcal{G}()$ to predict the exo masks and compute the loss $\mathcal{L}_{mask}$ as,
\begin{gather}
    \mathcal{L}_{mask} = \mathcal{G}(f_I^{'''}, E_{con}^{*}, E_M).
\end{gather}

\noindent \textbf{3) Training and Inference:}
Our model has two training stages:
a) \textbf{S1}: The MCFuse module is trained with the $\mathcal{L}_{mask}$ loss to initialize it well for the second stage.
b) \textbf{S2}: All modules (indicated by the fire icon) are trained jointly with 
the final loss $\mathcal{L} = \mathcal{L}_{mask}+\mathcal{L}_{X{obj}}$.
During inference, the XObjAlign module and also exo mask prompt $P_{vis}$ are removed, and the remaining model is used for inference.

\subsection{Multimodal Condition Fusion}~\label{sec:mcfuse}
Our Multimodal Condition Fusion (MCFuse) module is mainly presented to incorporate text descriptions into the model, serving as an additional prompt alongside visual cues to enhance the model's ability on locating target objects.
There are mainly two considerations for the design of MCFuse, 1) Given the effectiveness of cross-attention in capturing intricate relationships across modalities, it serves as an ideal choice for fusing our multimodal conditions; 2) the model-generated texts can sometimes be unreliable, prioritizing visual information is prudent. This motivates our use of a residual design, with the visual input serving as the primary pathway. Additionally, to eliminate the need for manual tuning of the residual pathway’s weight, we introduce a learnable parameter, $k_{lea}$.

\begin{figure}
    \centering
       {\includegraphics[width=0.8\linewidth]{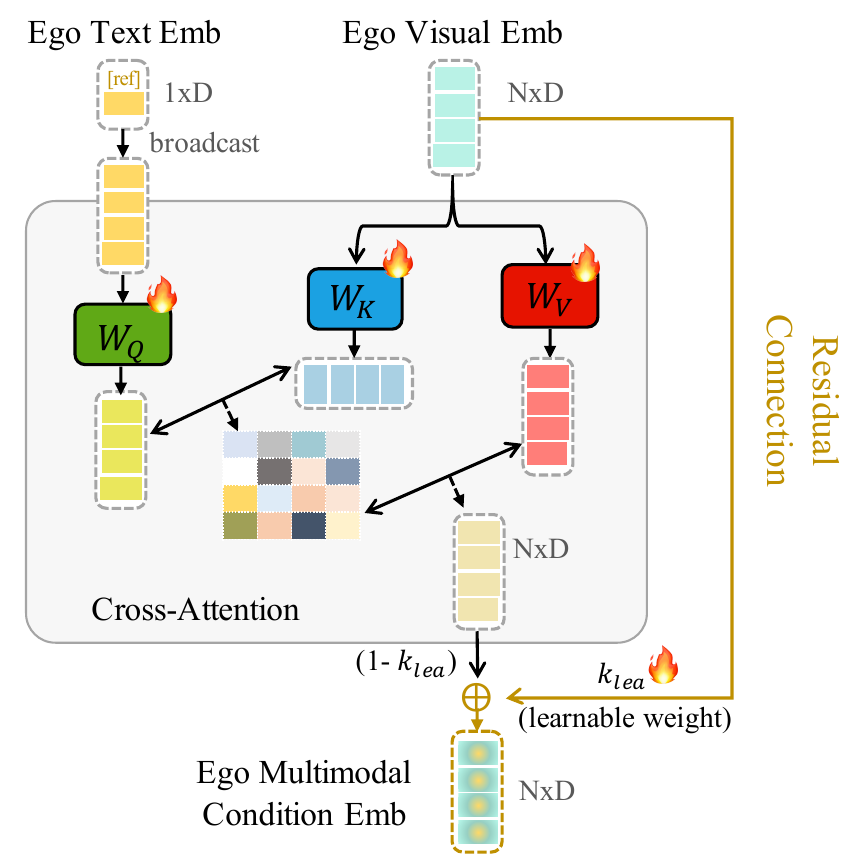}}
       \vspace{-0.1in}
    \caption{Architecture of our Multimodal Condition Fusion (MCFuse) module. All learnable sub-modules are denoted by fire icon.     \label{fig:mcfuse} }
    \vspace{-0.3in}
\end{figure}

Formally, as shown in Fig.~\ref{fig:mcfuse}, given the ego text emb $E_{txt}^{*} \in 1 \times D$ and ego visual emb $E_{vis}^{*} 
\in N\times D$ as input, the cross-attention is applied to the broadcasted $E_{txt}^{*}$ and $E_{vis}^{*}$, with $E_{txt}^{*}$ severs as query,  $E_{vis}^{*}$ works as key and value. The fused result $CA_{fuse}$ is given by: 
\begin{equation}
\textit{CA}_{\text{fuse}} = \operatorname{\textit{CrossAtt}}(E_{txt}^{*}W_Q, E_{vis}^{*}W_K, E_{vis}^{*}W_V),
\end{equation}~\label{eq:ca}
where \textit{CrossAtt()} means the cross attention, $W_{Q}$, $W_{K}$, $W_{V}$ denote the learnable parameters. 
After that, we combine the $CA_{fuse}$ with the initial $E_{vis}^{*}$ in a residual connection way with the learnable weight $k_{lea}$ as, 
\begin{equation}
    E_{con}^{*} = k_{lea} \cdot E_{vis}^{*} + (1-k_{lea}) \cdot CA_{fuse}.
\end{equation}

Although the cross-attention is not new, we highlight that: 1) We are the first to introduce text modality into this task and leverage it to enhance object localization. Results in Tab.~\ref{tab:abla-modules} and Fig.~\ref{fig:vis-modules} well show its effectiveness. 2) While prior segmentation models, including the base PSALM, allow various condition prompts, they do not utilize them simultaneously. In contrast, our approach integrates multimodal prompts to refer to the same object(s).

\subsection{Cross-View Object Alignment}
The Cross-View Object Alignment (XObjAlign) is mainly proposed to enhance the model's consistency across totally different views. The main insight behind this module is that a robust model should generate similar embeddings for both ego-masked and exo-masked object(s). Furthermore, since the ego visual embedding $E_{vis}^{*}$ serves as a conditioning factor, its quality directly affects the predictions of the Mask Generator.
Thus, we propose to apply the consistency constraint exactly in the \textit{object-level} embedding space. 

As in Sec.~\ref{sec:framework}, to achieve such an consistency constraint, we extract both the ego visual emb $E_{vis}^{*}$ and exo visual emb $E_{vis}$, after that, the only thing left for XObjAlign is to compare the similarities of these two embeddings and resulting in the cross-view object consistency loss $\mathcal{L}_{Xobj}$ as,
\begin{equation}
  \mathcal{L}_{Xobj} = \operatorname{\textit{Dist}}(E_{vis}^{*}, E_{vis}),
\end{equation}
where the \textit{Dist()} represents the Euclidean distance.
Without introducing additional parameters or supervision, our XObjAlign module ensures cross-view consistency, effectively handling object alignment across different viewpoints.

\section{Experiments}\label{sec:exp}
\noindent\textbf{Datasets.} 
We validate our method mainly on the large-scale Ego-Exo4D~\cite{grauman2024ego}, using its benchmark for ego-exo object correspondence tasks. The dataset contains 1.8 million annotated object masks at 1 fps across 1335 video takes, covering domains like Cooking, BikeRepair, Health, Music, Basketball, and Soccer. Each paired video features an average of 5.5 objects, tracked for an average of 173 frames.
We use the standard Train/Val/Test splits but make the following adjustments: 1) To focus on cross-view segmentation, we retain only data where objects appear in both views; 2) We introduce a ``Small" TrainSet (about one-third of the original ``Full" TrainSet) for improved efficiency and data storage; 3) Testing is conducted on the Val set due to the lack of ground truth for the Test set. Our processed splits will be made available for future comparisons. 
To evaluate methods more comprehensively, we introduce HANDAL-X, another cross-view object segmentation benchmark adapted from the HANDAL~\cite{handaliros23}. More details are in the Supplementary Materials (\textit{\textit{Supp. Mat.}})

\noindent\textbf{Network Modules.}
Following PSALM~\cite{zhang2024psalm}, we use Swin-B~\cite{liu2021Swin} as the Visual Encoder and Phi-1.5 1.3B~\cite{li2309textbooks} as the LLM. The MM Projector, consisting of Conv2d, BatchNorm2d, and FC layers, maps the visual latent space to the LLM space. The Pixel Decoder and Mask Generator are adapted from Mask2Former~\cite{cheng2022masked}. 
MCFuse is described in Sec.~\ref{sec:mcfuse}, and XObjAlign requires no parameters.

\noindent\textbf{Metrics.} 
Following standard segmentation tasks, we use IoU as the primary metric. We also report Location Error (LE), Contour Accuracy (CA), and Visibility Accuracy (VA) for main results on Ego-Exo4D as proposed in~\cite{grauman2024ego}. Briefly, LE measures the normalized distance between centroids of predicted and ground-truth masks, CA evaluates mask shape similarity, and VA assesses model's ability to estimate object existence visibility in the target view.

\noindent\textbf{Implementation Details.}
The pre-trained PSALM model serves as our initialization. In \textbf{S1} training, we use $^{1}\!/\!_{20}$
  of the training data to train MCFuse. In \textbf{S2} training, the full training set is used for optimizing all modules except the Swin-B vision encoder. 
Training details such as gpus, epochs, batch size, and learning rate are provided in the \textit{Supp. Mat}.

\begin{table*}[t]
    \centering 
    \setlength\tabcolsep{5pt}
    \scalebox{0.8}{
    \begin{tabular}{lcccccc|lcccccc}
    \toprule[0.95pt]
    \multicolumn{7}{c|}{\textbf{Ego2Exo (Ego as Query)}}       & \multicolumn{7}{c}{\textbf{Exo2Ego (Exo as Query)}}                                                       \\ \midrule[0.6pt]
    \textbf{Method}& \textbf{ZSL} & \textbf{TrainSet} & \textbf{IoU}$\uparrow$ & \textbf{LE}$\downarrow$ & \textbf{CA}$\uparrow$  & \textbf{VA}$\uparrow$ & \textbf{Method} & \textbf{ZSL} & \textbf{TrainSet} & \textbf{IoU}$\uparrow$ & \textbf{LE}$\downarrow$ & \textbf{CA}$\uparrow$  & \textbf{VA}$\uparrow$ \\ \midrule[0.6pt] 
    XSegTx$^\bullet$~\cite{grauman2024ego} &\ding{52} &  -  & 0.3   & 0.149 & 0.012 & 98.9 
    & XSegTx$^\bullet$~\cite{grauman2024ego}&\ding{52}  &  -  & 1.3  & 0.188 & 0.023 & 98.0\\    
    XSegTx$^\bullet$~\cite{grauman2024ego}&\ding{55} &  Full  & 6.2     & 0.091 & 0.171 & 96.5   
    & XSegTx$^\bullet$~\cite{grauman2024ego} & \ding{55} & Full  & 30.2 & 0.107 & 0.391 & 96.9 
   \\
    XView-Xmem$^\bullet$~\cite{grauman2024ego}&\ding{52} 
    & -  &   16.2     & 0.145 & 0.212 & 61.9  
    & XView-Xmem$^\bullet$~\cite{grauman2024ego} & \ding{52} & -  &   13.5  & 0.153 & 0.178 & 34.4
   \\
    XView-Xmem$^\bullet$~\cite{grauman2024ego}& \ding{55}   & Full  &   17.7   & 0.146 & 0.259 & 85.0           
    & XView-Xmem$^\bullet$~\cite{grauman2024ego} & \ding{55}  & Full  &   20.7   & 0.140 & 0.293 & 75.2
   \\
    Xmem+XSegTx$^\bullet$~\cite{grauman2024ego}& \ding{55} 
    & Full  &  36.9     & 0.068 & 0.447 & 85.6
    & Xmem+XSegTx$^\bullet$~\cite{grauman2024ego}& \ding{55}  & Full  &  36.1  & 0.117 & 0.438 & 96.0 \\
    \midrule[0.6pt]
    SEEM$^\bullet$~\cite{zou2024segment} &\ding{52}   &  -     &    1.1   & 0.146 & 0.052 & 99.3  &         
    SEEM$^\bullet$~\cite{zou2024segment} & \ding{52}   &  -      &    4.1  & 0.183 & 0.078 & 99.2
    \\
    PSALM$^\circ$~\cite{zhang2024psalm} & \ding{52}         & -     &    7.9    & 0.156 & 0.178 & 99.3        
    & PSALM$^\circ$~\cite{zhang2024psalm} & \ding{52}     & -      &    9.6  & 0.140 & 0.177 & 99.3 
    \\
    PSALM$^\bullet$~\cite{zhang2024psalm} & \ding{55}    & Small     &    39.7   & 0.058 & 0.537 & 99.8 
    & PSALM$^\bullet$~\cite{zhang2024psalm} & \ding{55}  & Small     &      44.1  & 0.067 & 0.553 & 99.7
    \\
    PSALM$^\bullet$~\cite{zhang2024psalm} & \ding{55}  & Full      &  41.3   & 0.065 & 0.536 & 99.8
    & PSALM$^\bullet$~\cite{zhang2024psalm} & \ding{55}  & Full      &     47.3   & 0.067 & 0.571 & 99.7
    \\ \midrule[0.6pt]
    \cellcolor{cyan!10}\textbf{ObjectRelator (Ours)}&\cellcolor{cyan!10}\ding{55}& \cellcolor{cyan!10}Small                 &   \cellcolor{cyan!10}\textbf{44.3} & \cellcolor{cyan!10}\textbf{0.051} & \cellcolor{cyan!10}\textbf{0.546} & \cellcolor{cyan!10}\textbf{99.9}
    &\cellcolor{cyan!10}\textbf{ObjectRelator (Ours)}& \cellcolor{cyan!10}\ding{55}    & \cellcolor{cyan!10}Small                 &     \cellcolor{cyan!10}\textbf{49.2}  & \cellcolor{cyan!10}\textbf{0.065} & \cellcolor{cyan!10}\textbf{0.577} & \cellcolor{cyan!10}\textbf{99.8} \\
    \cellcolor{cyan!10}\textbf{ObjectRelator (Ours)}&\cellcolor{cyan!10}\ding{55}    & \cellcolor{cyan!10}Full                  &         \cellcolor{cyan!10}\textbf{45.4}& \cellcolor{cyan!10}\textbf{0.056} &  \cellcolor{cyan!10}\textbf{0.548} & \cellcolor{cyan!10}\textbf{99.9} 
    & \cellcolor{cyan!10}\textbf{ObjectRelator (Ours)}& \cellcolor{cyan!10}\ding{55}      &  \cellcolor{cyan!10}Full &     \cellcolor{cyan!10}\textbf{50.9}   & \cellcolor{cyan!10}\textbf{0.064} & \cellcolor{cyan!10}\textbf{0.586} & \cellcolor{cyan!10}\textbf{99.8}
    \\ \bottomrule[0.95pt]
    \end{tabular}
    }
    \caption{Comparison results on Ego-Exo4D object correspondence benchmark. All the results are tested on the Val set. $\circ$ means results from PSALM~\cite{zhang2024psalm}, $\bullet$ means results are reported by us. We use ``Xmem+XSegTx" for short of ``XView-Xmem+XSegTx".
    \vspace{-0.1in}
    } 
    \label{tab:main-egoexo}
\end{table*}

\subsection{Main Results on Ego-Exo4D} 
\noindent\textbf{Baselines and Competitors.} 
Directly aligned prior methods are quite limited, while still some efforts have been made. Specifically, 1) Ego-Exo4D~\cite{grauman2024ego} forms a spatial-based ``XSegTx" method by adapting SegSwap~\cite{shen2021learning}, and two spatiotemporal-based ``XView-XMem", ``XView-XMem+XSegTx" via adapting XMem~\cite{cheng2022xmem} and combing it with ``XSegTx"; 2) PSALM~\cite{zhang2024psalm} tests its ZSL results; 3) 
In this paper, we also adapt the SEEM~\cite{zou2024segment}, one additional universal segmentation method that decodes visual prompts by referring to the image. The adaptation is mainly made by using the query view (e.g., ego in Ego2Exo) to generate the prompt for the target view (e.g., exo in Ego2Exo). 4) Additionally, PSALM is retrained as a competitive baseline.

\noindent\textbf{Results on Ego-Exo4D.} The comparison results are summarized in Tab.~\ref{tab:main-egoexo}. 
We indicate whether the method employs ZSL learning, the TrainSet used, and several metrics: IoU, LE, CA, and VA.
From the results, we first notice that our proposed ObjectRelator model significantly improves the base PSALM and achieves state-of-the-art performance on the ego-exo object correspondence task. 
Take models trained on Small TrainSet as an example, our ObjectRelator improves the PSALM from 39.7 to 44.3 on Ego2Exo and from 44.1 to 49.2 on Exo2Ego. 
Regarding LE, CA, and VA, Our ObjectRelator consistently outperforms PSALM, achieving clear margins over all other competitors.
These results well demonstrate the effectiveness of our approach. 

In addition to our SOTA results, several other key observations include:
1) \textbf{Different Metrics:} Although IoU, LE, and CA are defined differently, they all fundamentally measure the discrepancy between the predicted and ground truth masks. Our results also show high consistency among them. Additionally, most of the models exceed 96\% on VA, indicating that object visibility estimation is relatively easy. Thus, we mainly use IoU for further analysis.
2) \textbf{ZSL Inference Performance:} 
Results show that nearly all the tested methods, including universal segmentation models like SEEM and PSALM, struggle to generalize smoothly to this task.
While SEEM and PSALM excel on other out-of-domain tasks, such as open-vocabulary segmentation~\cite{ding2022open} and video object tracking~\cite{pont20172017}, their performances are degraded here. 
This underscores that simply applying existing models, even those extensively pre-trained ones, is insufficient. Dedicated models specifically designed and trained for cross-view scenarios remain essential. XView-XMem shows the best ZSL performance, likely due to its video object tracking foundation. 
However, this advantage diminishes in the retrained XView-XMem.
3) \textbf{Baseline Retraining:} After retraining models on the benchmark, all baselines show reasonable performance, further emphasizing the importance of model adaptation for this task. In general, the superior performance of XView-Xmem and ``Xmem+XSegTx" over XSegTx suggests that temporal information aids cross-view alignment. However, this advantage can be offset by the model designs, as demonstrated by PSALM. 
The success of PSALM makes it a strong reason for choosing it as our baseline.
4) \textbf{Differences in Query Types:} Across models, performance is in general better on Exo2Ego than Ego2Exo, 
indicating that ego-view queries are more challenging. 
XSegTx exemplifies this disparity, achieving an IoU of 30.2 on Exo2Ego but only 6.2 on Ego2Exo. This aligns with intuitive expectations, as exo-view objects tend to be smaller and embedded within complex environments, increasing segmentation difficulty, while ego-view objects appear larger and more distinguishable.
5) \textbf{Impact of Train Set Size:} 
Models trained on the Full TrainSet generally perform better, but the Small TrainSet also yields comparable results.
This demonstrates that the Small TrainSet is a practical alternative for benchmarking this task while reducing data requirements.

\subsection{Ablation on Proposed Modules} 
To evaluate the contribution of each component, starting with the base PSALM, we incrementally add the MCFuse and XObjAlign to observe their individual impact. Results are reported in Tab.~\ref{tab:abla-modules}, in addition to the IoUs, the numbers of model parameters (No.Param.) are also provided.

\begin{table}[h]
    \centering 
    \setlength\tabcolsep{1.5pt}
    \setlength{\extrarowheight}{0.5pt}
    \scalebox{0.77}{
    \begin{tabular}{l|cccc|l}
    \toprule[0.95pt]
    \textbf{Method} & \textbf{MCFuse} & \textbf{XObjAlign} &  \textbf{Ego2Exo}$\uparrow$& \textbf{Exo2Ego}$\uparrow$ & \textbf{No.Param.} \\ \midrule[0.6pt]
    Base & \ding{55}  &  \ding{55}  &  39.7 &   44.1  &  1.5871B \\  \hline
    +MCFuse   &  \ding{52} &  \ding{55}  &  43.2 &   47.4  & + 0.2632M \\  \hline
    +XObjAlign&  \ding{55} &  \ding{52}  &  43.8 &   48.3  & + 0 \\ \hline
    \cellcolor{cyan!10}\textbf{ObjectRelator}   &  \cellcolor{cyan!10}\ding{52}  &  \cellcolor{cyan!10}\ding{52}  &  \cellcolor{cyan!10}\textbf{44.3}   & \cellcolor{cyan!10}\textbf{49.2}  & \cellcolor{cyan!10} 1.5873B\\
    \bottomrule[0.95pt]
    \end{tabular}
    }
    \vspace{-0.05in}
    \caption{Effectiveness of proposed modules. IoUs are reported. Methods trained on Small TrainSets. M/B means million/billion.}
    \label{tab:abla-modules}
    \vspace{-0.1in}
\end{table}

Results show that: 1) Incorporating MCFuse into the base model significantly boosts performance, with clear improvements of 3.5 and 3.3 on Ego2Exo and Exo2Ego, respectively. This strongly supports our idea that introducing the language condition and effectively fusing it with the visual mask condition helps better identify the target object, particularly by reducing ambiguities.
2) Applying our XObjAlign module alone also brings significant improvements, boosting the base model by 4.1 and 4.2 on the two subtasks, surpassing even the benefits from MCFuse with its simpler design. These results highlight that narrowing the gap between ego and exo views is a key step toward building a more robust predictor. 
In our paper, we address or at least mitigate this gap by enforcing cross-view object consistency.
3) By integrating these two modules, our full ObjectRelator approach achieves the best results, indicating that the two modules are effectively compatible. However, we also observe that the combined effect is not simply additive. We interpret this as an overlap in the optimization directions of the two modules. For example, when the XObjAlign totally aligns the ego-exo views, the likelihood of the model misidentifying the objects across views is already low.
4) Another point worth mentioning is that our method in total only introduces 0.2632M extra parameters, which is around 0.016$\%$ of the base PSALM while achieving 11.6$\%$ improvement on average.

More ablations, including multimodal fusion strategies, the impact of the learnable residual connection, comparisons of learnable $k_{lea}$ vs. fixed ratio, and different loss weightings between $\mathcal{L}_{mask}$, $\mathcal{L}_{X{obj}}$, are in the \textit{Supp. Mat}.

\subsection{More Analysis on Ego-Exo4D}

\noindent\textbf{Joint Training of Ego2Exo and Exo2Ego.} 
All previously reported results are obtained by training the Ego2Exo and Exo2Ego models separately, using their respective training sets. Thus, in this experiment, we investigate whether the training process can be merged by training a single model on both the Ego2Exo and Exo2Ego datasets simultaneously. The comparative results, shown in Tab.~\ref{tab:analysis-joint}, confirm that the answer is yes. Our joint-trained models don't degrade the performance of the separately trained models in most cases and even outperform the specific models. This suggests that the bidirectional correspondence learned during co-training helps the model understand the relationship between ego and exo perspectives more comprehensively.

\begin{table}[h]
    \centering
    \setlength\tabcolsep{2pt}
    \setlength{\extrarowheight}{0.5pt}
    \scalebox{0.70}{
    \begin{tabular}{l|c|c|cc}
    \toprule[0.95pt]
    \textbf{Method}     & \textbf{TrainSet}    & \textbf{Training} & \textbf{Ego2Exo Testing}$\uparrow$ & \textbf{Exo2Ego Testing}$\uparrow$ \\ \midrule[0.6pt]
    \textbf{ObjectRelator} & Small & Ego2Exo           &     44.3              &      -            \\ \cline{3-5} 
                             & & Exo2Ego           &        -           &          49.2        \\ \cline{3-5} 
                             & & \textbf{Joint Training}        &        \cellcolor{cyan!10}\textbf{44.7}         &        \cellcolor{cyan!10}\textbf{50.6}         \\ \cline{2-5} 
    
                             & Full & Ego2Exo           &     45.4              &      -            \\ \cline{3-5} 
                             & & Exo2Ego           &        -           &    \cellcolor{cyan!10}\textbf{50.9}              \\ \cline{3-5} 
                             & & \textbf{Joint Training}        &        \cellcolor{cyan!10}\textbf{46.7}         &        50.8         \\ \bottomrule[0.95pt]
    \end{tabular}
    }
    \vspace{-0.05in}
    \caption{Results of Joint Training on Small and Full TrainSets.}
    \vspace{-0.1in}
    \label{tab:analysis-joint}
\end{table}

\noindent\textbf{Single-Condition Evaluation of Multi-Condition Model.}
Our proposed ObjectRelator is trained and tested with both visual mask and text description conditions. However, since text descriptions might not always be available, we are curious about what would happen if we test our multi-condition trained model using only the visual mask condition. To this end, we evaluate the performance of removing the MCFuse module from our trained model, studying both the specific Ego2Exo and Exo2Ego models, as well as the joint-trained model. Results are summarized in Tab.~\ref{tab:analysis-singlecond}. 

\begin{table}[h]
    \centering
    \setlength\tabcolsep{2.3pt}
    \scalebox{0.66}{
    \begin{tabular}{l|c|c|cc}
    \toprule[0.95pt]
    \textbf{Method}                & \textbf{Training}         & \textbf{Testing Conditions} & \textbf{Ego2Exo Testing}$\uparrow$ & \textbf{Exo2Ego Testing}$\uparrow$ \\ \midrule[0.6pt]
    \textbf{PSALM}~\cite{zhang2024psalm} & Specific & Only Visual               & 39.7                   & 44.1                    \\ \midrule[0.6pt]
    \multirow{4}{*}{\textbf{ObjectRelator}} & \multirow{2}{*}{Specific} & Visual + Text               & \cellcolor{cyan!10}\textbf{44.3}                     & \cellcolor{cyan!10}\textbf{49.2}                     \\ \cline{3-5}
                                   &                           & Only Visual                 &     43.3                     &    48.9                      \\ \cline{2-5}
                                   & \multirow{2}{*}{Joint}    & Visual + Text               & \cellcolor{cyan!10}\textbf{44.7}                     & \cellcolor{cyan!10}\textbf{50.6}                     \\ \cline{3-5}
                                   &                           & Only Visual                 & 44.0                     & 50.2  \\
    \bottomrule[0.95pt]
    \end{tabular}
    }
     \vspace{-0.05in}
    \caption{Results of single condition testing. The studied Specific and Joint models are trained on Small TrainSets.}
    \vspace{-0.15in}
    \label{tab:analysis-singlecond}
\end{table}

The testing results show that while the performance with only the visual condition is inferior to using both visual and text conditions, the drop in performance is quite slight, and we still outperform the base PSALM model.  This outcome exceeds our expectations, considering the inconsistency between training and testing conditions. Our interpretation is that, although we are exploring multimodal fusion rather than multimodal alignment, the model still manages to learn a joint space between the two different modality conditions. This enables the model to associate object categories with masks, allowing it to still perform well during testing, even when in the absence of text information.

\subsection{Main Results on HANDAL-X}
To further validate our method, we conduct experiments on HANDAL-X. The comparison results, summarized in Tab.~\ref{tab:handal}, include XSegTx, SEEM, and PSALM as competitors. Note that "XView-XMem" and "XMem+XSegTx" are excluded since HANDAL-X is a cross-view \textit{image} segmentation dataset. To differentiate models using different checkpoints, we also indicate the primary datasets used.

\begin{table}[!t]
    \centering
    \scalebox{0.7}{
    \begin{tabular}{l|l|l|l}
    \toprule[0.95pt] 
    \textbf{Setting} & \textbf{Method} &  \textbf{Datasets} & \textbf{IoU}$\uparrow$  \\ 
    \midrule[0.6pt]
    \textbf{ZSL} 
    & XSegTx$^\circ$~\cite{grauman2024ego} & Self-Generated Pairs from COCO & 1.5 \\
    & SEEM$^\circ$~\cite{zou2024segment} &  COCO Panoptic, RefCOCO/+/g & 2.5 \\
    & PSALM$^\circ$~\cite{zhang2024psalm} & COCO Panoptic, RefCOCO/+/g, ... & 14.2 \\
    \cline{2-4}
    & PSALM$^\bullet$~\cite{zhang2024psalm} & Ego-Exo4D & 39.9 \\
    & \cellcolor{cyan!10}\textbf{ObjectRelator}$^\bullet$ & \cellcolor{cyan!10}Ego-Exo4D & \cellcolor{cyan!10}\textbf{42.8} \\
    \midrule[0.6pt]
    \textbf{Retrained} & 
    PSALM$^\bullet$~\cite{zhang2024psalm} & HANDAL-X & 83.4 \\
    & \cellcolor{cyan!10}\textbf{ObjectRelator}$^\bullet$ & \cellcolor{cyan!10}HANDAL-X & \cellcolor{cyan!10}\textbf{84.7} \\
    \bottomrule[0.95pt]  
    \end{tabular}
    }
    \vspace{-0.05in}
    \caption{Comparison results on HANDAL-X benchmark. $\circ$ means checkpoints from priors works; $\bullet$ means models trained in this paper; For the PSALM and ObjectRelator trained on Ego-Exo4D, the checkpoints with Exo2Ego Small TrainSets are used. }
    \vspace{-0.15in}
    \label{tab:handal}
\end{table}

We observe that:
1) This benchmark is generally easier than Ego-Exo4D, where extreme viewpoint changes are considered. However, it also serves as a valuable testbed for more common scenarios.
2) ZSL results follow a similar trend to Ego-Exo4D, with PSALM outperforming others, followed by SEEM.
3) A key finding is that models trained on Ego-Exo4D (PSALM and ObjectRelator) demonstrate strong ZSL ability, significantly surpassing those trained on other datasets (e.g., COCO) with an IoU of around 40. This suggests that training on cross-view data improves generalization to new cross-view settings, even with different scenarios and categories.
4) Across both ZSL and retrained settings, ObjectRelator consistently outperforms PSALM. This further validates our approach.

We highlight that although HANDAL-X is relatively simple, it remains valuable: 1) it effectively reveals performance differences (ranging from 1.5 to 84.7); 2) it serves as a testbed for zero-shot learning; and 3) its natural characteristics—robotics-ready, manipulable objects (e.g., a spoon with appropriate size and shape)—make it particularly suitable for robotics-related applications.

\subsection{Qualitative Results}
\begin{figure}[h]
    \centering
       {\includegraphics[width=1.\linewidth]{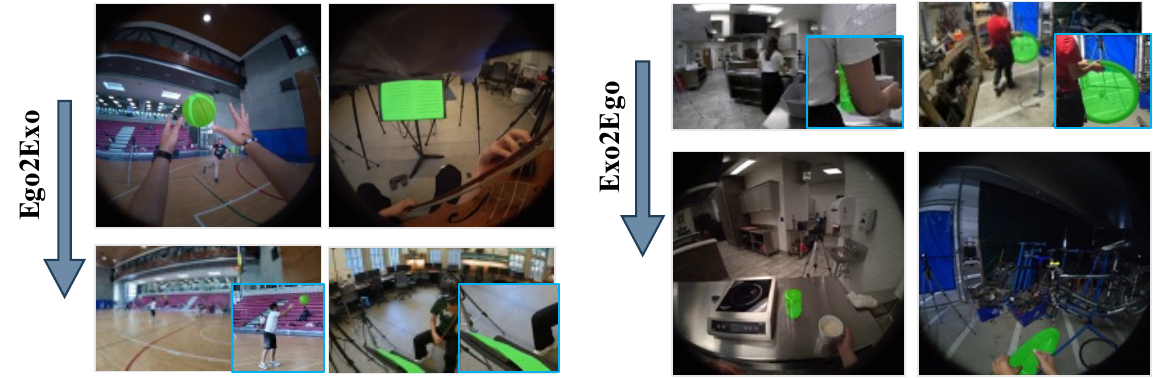}}
    \caption{ObjectRelator Visualization for Ego2Exo and Exo2Ego.}
     \vspace{-0.05in}
    \label{fig:vis-overall}
    \vspace{-0.1in}
\end{figure}

To provide a more intuitive understanding of our method, we visualize the prediction results. Specifically, Fig.~\ref{fig:vis-overall} presents several examples of ObjectRelator applied to both Ego2Exo and Exo2Ego tasks. The results indicate that our proposed methods perform effectively across diverse scenarios e.g., Basketball, Music, Cooking, and Bike Repair. Our method successfully segments objects in the target view despite significant size changes, shape variations, and cases where the query object is partially occluded.

\begin{figure}[t]
    \centering
       {\includegraphics[width=1.\linewidth]{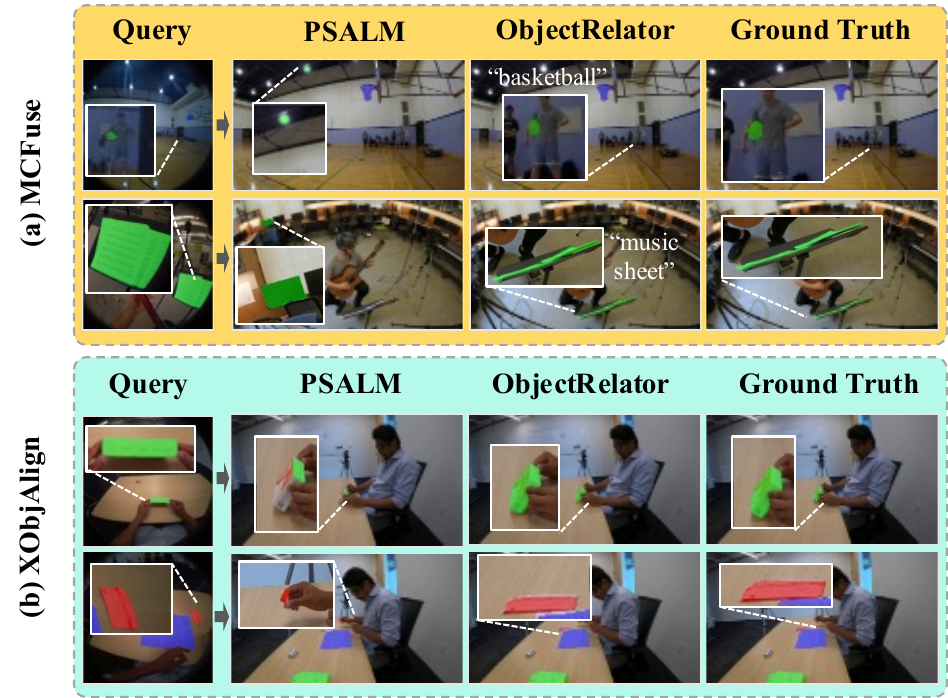}}
    \vspace{-0.1in}
    \caption{ObjectRelator vs. PSALM Visualization Results.}
    \label{fig:vis-modules}
     \vspace{-0.2in}
\end{figure}

To further demonstrate how our proposed modules improve upon the base PSALM, Fig.~\ref{fig:vis-modules} compares the results between ObjectRelator and the retrained PSALM model, with ground truth provided as a reference due to the challenging nature of the examples. In Fig.~\ref{fig:vis-modules}(a), PSALM shows that PSALM will segment the wrong object with a similar shape to the query, as seen in the basketball and music sheet examples. This indicates that relying solely on a visual mask fails to capture necessary semantic information. By contrast, our ObjectRelator, which leverages text descriptions to enhance the original visual mask condition, successfully corrects these errors in both examples, underscoring the effectiveness of our MCFuse module.
Fig.~\ref{fig:vis-modules} (b) presents examples where PSALM either segments only part of the target object (first example) or segments an incorrect object (second example). In both cases, the partially or incorrectly segmented object closely resembles the query object, suggesting that PSALM performs reasonably well but falls short in meeting the specific requirements of the ego-exo object correspondence task, where the same object may appear very different across views. The success of our method demonstrates the effectiveness of our proposed XObjAlign, which enforces consistency between ego and exo views as a solution.
More visualization results on Ego-Exo4D and HANDAL-X can be found in the \textit{Supp. Mat}.

\section{Conclusion}
In this paper, we explored the Ego-Exo Object Correspondence task, a newly proposed and underdeveloped vision challenge. We introduced the ObjectRelator method, which incorporates two novel modules: Multimodal Condition Fusion (MCFuse) and SSL-based Cross-View Object Alignment (XObjAlign). MCFuse facilitates multi-condition fusion between language and visual modalities, while XObjAlign enforces consistency in object representation across different views through a self-supervised alignment strategy. Extensive experimental results demonstrate the effectiveness of our approach, showing significant improvements over the baseline and achieving state-of-the-art performance. As an early exploration of this task, we hope our work sparks further interest in this direction and paves the way for future research in comprehensive cross-view object understanding applications.

\clearpage

\section{Acknowledgments}
This research was partially funded by the Ministry of Education and Science of Bulgaria (support for INSAIT, part of the Bulgarian National Roadmap for Research Infrastructure). This project was supported with computational resources provided by Google Cloud Platform (GCP).
{
    \small
    \bibliographystyle{ieeenat_fullname}
    \bibliography{main}
}

\input{X_suppl}

\end{document}

%% file: X_suppl.tex
\clearpage
\setcounter{page}{1}

\maketitlesupplementary
\setcounter{section}{0}
\setcounter{figure}{0}    
\setcounter{table}{0}    

\renewcommand{\thetable}{\Alph{table}}
\renewcommand{\thefigure}{\Alph{figure}}

We first present the implementation details in Sec.~\ref{sec:supp-imple}, followed by additional results in Sec.~\ref{sec:supp-result}, Sec.~\ref{sec:supp-fail}, and conclude with a discussion of limitations and future work in Sec.~\ref{sec:supp-diss}.

\section{Implementation Details}\label{sec:supp-imple}
\subsection{Data Processing on Ego-Exo4D}
\noindent\textbf{Frame Extraction.} We follow the same frame extraction process as the baselines, i.e., XSegTx, XView-Xmem, as provided by Ego-Exo4D~\cite{grauman2024ego}. Specifically, 
for both ego and exo views, we sample one frame every 30 frames in chronological order to ensure time synchronization. 
Meanwhile, since the resolution of ego and exo is different, we adopt different scaling ratios. For ego video, we scale its resolution from (1408, 1408) to (704, 704). While the exo video is scaled from (2160, 3840) to (540, 960).

\noindent\textbf{Train/Val/Test Sets.}  
As described in Sec.~\ref{sec:exp}, we follow the same dataset splits as Ego-Exo4D~\cite{grauman2024ego}. By default, we retain only the ego-exo pairs where the object is visible in both views (in some cases, an object may be visible in one view but fully occluded in the other). When it comes to evaluating the VA, we test models on all cases.
To construct the \textbf{SmallTrain} set, we sample a subset of the \textbf{FullTrain} set at a fixed frequency of 1/3.
In the end, for both Ego2Exo and Exo2Ego tasks, we have the \textbf{FullTrain}, \textbf{SmallTrain}, and \textbf{Val} sets. We further report the number of frame images (No. Img) and the average number of objects per frame (Avg. Obj) for each set in Tab.~\ref{tab:supp-tab1}. All splits used in this work will be released to the community to facilitate reproduction and comparison.

\begin{table}[h]
    \centering
\resizebox{0.48\textwidth}{!}{
\begin{tabular}{l|ccc|ccc}
\toprule
                 & \multicolumn{3}{c|}{\textbf{Ego2Exo}} & \multicolumn{3}{c}{\textbf{Exo2Ego}} \\
\hline
                 & \textbf{FullTrain}    & \textbf{SmallTrain}   & \textbf{Val}   & \textbf{FullTrain}    & \textbf{SmallTrain}    & \textbf{Val}   \\
\hline
\textbf{No. Img} &  110118            &      36706         &  31205     &     123381         &       41127        &   36073    \\
\hline
\textbf{Avg. Obj}   &    2.4          &     2.4          &   2.4    &  2.3            &       2.3        &  2.3  \\  
\bottomrule
\end{tabular}
}
\caption{Number of images and average object per image of sets.}
\label{tab:supp-tab1}
\end{table}

\subsection{Data Processing on HANDAL-X}\label{sec:supp-handal}
We adapt the HANDAL-X from HANDAL~\cite{handaliros23} as a new cross-view image segmentation benchmark. The vanilla HANDAL dataset is designed for category-level object pose estimation and affordance prediction, specifically tailored for robotics applications. It emphasizes manipulable objects that are ideal for robot manipulators to grasp, such as pliers, utensils, and screwdrivers. The dataset comprises 308,000 annotated image frames extracted from 2,200 videos featuring 212 distinct objects categorized into 17 groups.

The HANDAL dataset provides multi-view images capturing objects from a full 360-degree perspective, along with object-centered masks, enabling the construction of a cross-view dataset. Specifically, HANDAL defines a training and testing split for each object, and we adopt the same partitioning strategy for HANDAL-X. To enhance viewpoint variation, we construct query-target pairs at 100-frame intervals within each object's training set. The same approach is applied to the test set. We then aggregate the training pairs from all objects to form the final training set, and similarly, we merge all test pairs to create the final test set. As a result, our final HANDAL-X dataset consists of 44,202 training samples and 14,074 test samples.

\subsection{Description Generation}
\noindent\textbf{Using LLaVA for Object Descriptions.}
Due to the lack of semantic information in the initial benchmark, we propose to utilize the pretrained vision language model LLaVA~\cite{liu2023llava} for generating the text description to support the subsequence MCFuse module. Concretely, we merge the mask of the object of interest with the correspondence image to create the input. 
This approach not only directs the model's attention to the masked area but also preserves the full contextual information, such as the background, which aids the model in better understanding the object.
As for the text prompt, we use 
\textit{``Identify the single object covered by the green mask without describing it. Note that it is not a hand. Format your answer as follows: The object covered by the green mask is''.} 
By providing this text prompt and the masked image to LLaVA, we generate the corresponding object descriptions.

\begin{figure*}[t!]
    \centering
       {\includegraphics[width=.85\linewidth]{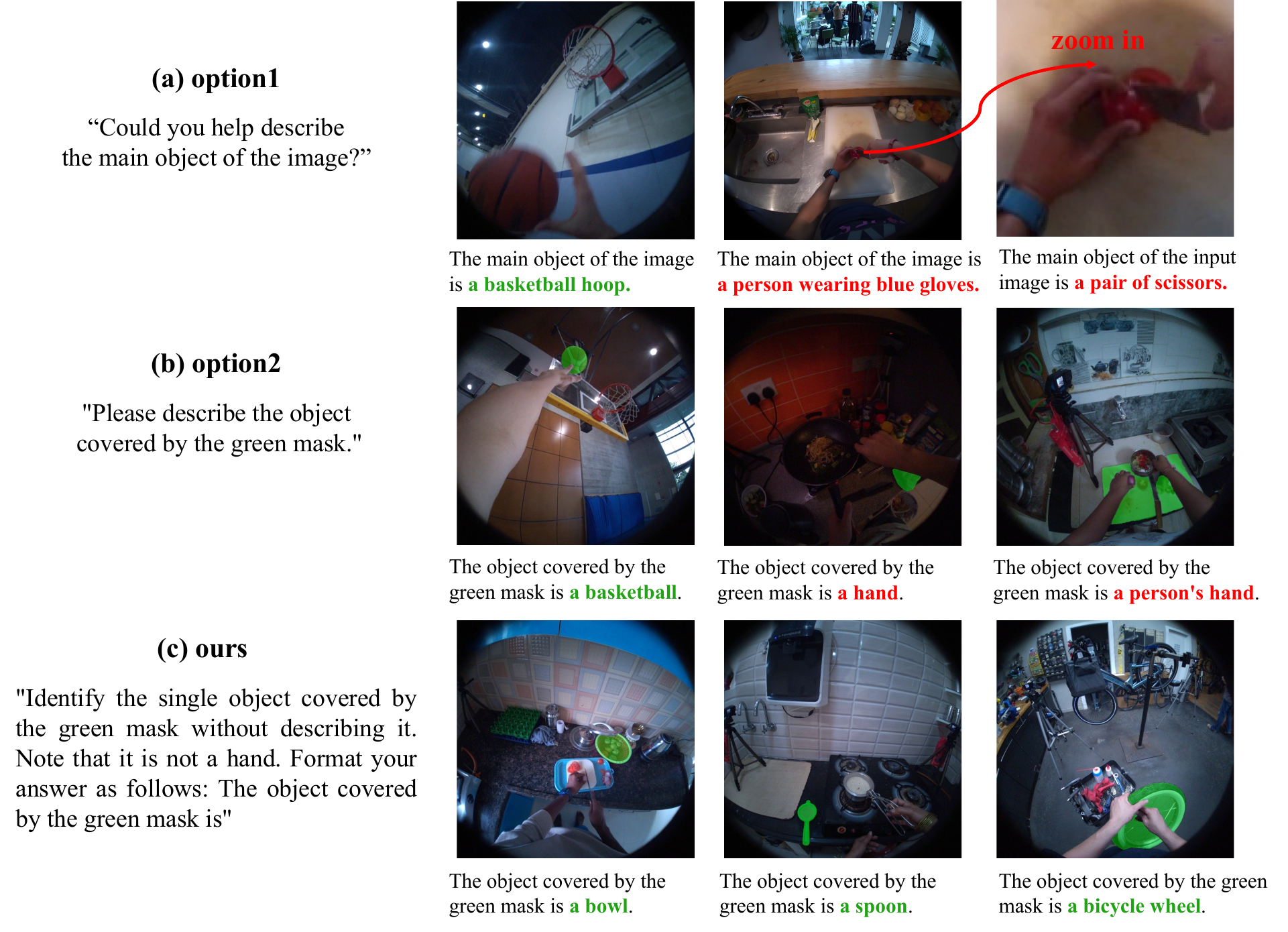}}
    \caption{Comparisons among various options for text generation.}
    \label{fig:vis-supp1}
\end{figure*}

\noindent\textbf{Comparison Among Different Options.} We compare our way of generating text descriptions with two other options. 
Visual examples are provided in Fig.~\ref{fig:vis-supp1}. 
Specifically:
1) \textbf{Without Object Masking (Fig.~\ref{fig:vis-supp1}(a)):} The entire image is used as input without adding an object mask, and the prompt is: \textit{``Could you help describe the main object of the image?"}. Results show that when the image contains multiple objects, the model struggles to identify the correct one. Attempts to zoom in on the target object lead to reduced resolution, making the image difficult for the model to interpret.
2) \textbf{With Object Masking but different prompt (Fig.~\ref{fig:vis-supp1}(b)):} The object is masked, and the model is prompted with \textit{``Please describe the object covered by the green mask."} While this approach focuses the model's attention on the masked area, it often incorrectly predicts the object as a human hand, which is not considered an object of interest in this benchmark.
3) \textbf{Our Approach (Fig.~\ref{fig:vis-supp1}(c)):} We use an object-masked image as input and a prompt that explicitly excludes the human hand. Results show significant improvement, as the generated descriptions align with the target object categories in most cases.

\noindent\textbf{Error Rate and Ambiguity Robustness Analysis.}
To further quantitatively assess the quality of our generated text descriptions, we randomly sample 50 frames from each scenario and evaluate both error rate and ambiguity robustness. For the error rate, a description is considered correct if it accurately reflects the object in the image; otherwise, it is counted as an error. We report results using both human and GPT-4o as judges. For ambiguity robustness, we run inference using our trained model on the subset of samples where the descriptions were judged correct by humans. We then report the IoU scores and compare them with those obtained using ground-truth text as input. Results are summarized in Tab.~\ref{tab:abla-text}.

\begin{table}[h]
    \centering
    \scalebox{0.79}{
    \begin{tabular}{l|ll}
    \toprule[0.95pt] 
    \textbf{Evaluation Type} & \textbf{Method} & \textbf{Metrics} \\
    \midrule[0.6pt]
    \textbf{Error Rate}     & Human Judge   & 22.1\% \\
                & GPT-4o Judge& 16.9\% \\
    \midrule[0.6pt]
    \textbf{Ambiguity Robust} & Our predicted text                     & 54.45 IoU \\
                & Ego-Exo4D GT text & 55.44 IoU \\
    \bottomrule[0.95pt]  
    \end{tabular}
    }
    \caption{Study on the quantity of generated text descriptions.}
    \vspace{-0.15in}
    \label{tab:abla-text}
\end{table}

From the results, we acknowledge that the generated descriptions are not always perfect, with an accuracy around 80\%. Given the complexity of ego-centric frames, such as small objects and motion blur, the overall quality remains strong. As shown in Tab.~\ref{tab:main-egoexo} and Tab.~\ref{tab:abla-modules}, the auto-generated text descriptions significantly contribute to performance improvements. Furthermore, the ambiguity robustness results indicate that our MCFuse module performs well.

\subsection{Training Details}
Almost all models, including the base retrained PSALM, our ObjectRelator, and the ablation studies of ObjectRelator, share the same basic training setup: the pretrained PSALM model~\cite{zhang2024psalm} is used as the initialization, AdamW is employed as the optimizer, the learning rate is select from {6e-5, 2e-4}  with a cosine decay scheduler, the batch size is 12, and the image size is 1024×1024. 
By default, models without MCFuse are trained in a single stage (4 epochs) using all the data in the current training set. In contrast, models with MCFuse, such as our ObjectRelator, employ two training stages, \textbf{S1} and \textbf{S2}, with different epoch settings and data usage. The specific training configurations for our ObjectRelator are summarized in Tab.~\ref{tab:supp-tab2}.

\begin{table}[h]
\centering
\resizebox{0.48\textwidth}{!}{
\begin{tabular}{l|c|c}
\toprule
            & \textbf{Ego2Exo/Exo2Ego/HANDAL-X Training}   & \textbf{Joint Training} \\
\hline
\textbf{S1} & $^{1}\!/\!_{20}$ set, epoch = 4 &   $^{1}\!/\!_{20}$ set, epoch = 3                        \\ \hline
\textbf{S2} & all set, epoch = 4                     &  all set, epoch = 3                   \\         
\bottomrule
\end{tabular}}
\caption{Training settings for ObjectRelator.}
\label{tab:supp-tab2}
\vspace{-0.1in}
\end{table}

All training runs on 4xA100 GPUs, while testing is conducted on a single A100, A6000, or L4. Training on Ego-Exo4D Small TrainSet or HANDAL-X takes 5-6 hours, while Ego-Exo4D Full TrainSet takes around 20 hours.

\section{More Results}\label{sec:supp-result}
\subsection{More Ablations on MCFuse}
To thoroughly assess the optimality of our design, we conduct additional ablations applying various options to MCFuse. These include testing: whether cross attention (CA) is the best method for fusing conditions; the impact of adopting a learnable residual connection; the usage of the learnable weight; and the most suitable placement for applying MCFuse. The results are presented in Tab.~\ref{tab:abla-exps}.

\begin{table}[h]
    \centering
    \scalebox{0.79}{
    \begin{tabular}{l|ll}
    \toprule[0.95pt] 
    \textbf{Ablation Factor} & \textbf{Method}    & \textbf{IoU}$\uparrow$  \\ 
    \midrule[0.6pt]
    \textbf{Fusion}      & Add   & 42.6 \\
                & CA w/o Params & 42.1 \\
                & CA, S2           & 22.0    \\
                & CA+LearnResidual, S2 & 41.3 \\
                & \cellcolor{cyan!10}CA+LearnResidual, S1-2 (Ours)      & \cellcolor{cyan!10}\textbf{43.2} \\
    \midrule[0.6pt]
    \textbf{Residual Weight} & $k_{lea}=0.2$                     & 42.1 \\
                & $k_{lea}=0.5$                        & 41.7 \\
                & $k_{lea}=0.8$                        & 43.1 \\
                & \cellcolor{cyan!10}$k_{lea}$ (Ours) & \cellcolor{cyan!10}\textbf{43.2}  \\
    \midrule[0.6pt]
    \textbf{Placement of MCFuse}    
                & Before XObjAlign       & 43.6 \\
               & \cellcolor{cyan!10}After XObjAlign (Ours)          & \cellcolor{cyan!10}\textbf{44.3} \\
    \bottomrule[0.95pt]  
    \end{tabular}
    }
    \caption{Ablation study on MCFuse. Models are trained on Ego2Exo Small TrainSet.}
    \vspace{-0.05in}
    \label{tab:abla-exps}
\end{table}

For the \textbf{fusion}-related experiments, we designed four variants:  ``Add", ``CA w/o Params", ``CA, S2", ``CA + LearnResidual, S2". The ``Add" method simply adds the ego text and visual conditions as an easy test of our generated text descriptions and also the idea of multi-condition fusion. The ``CA w/o Params" variant applies cross-attention without parameters, formulated as $(E^{ego}_{T} \cdot {E^{ego}_{V}}^{T}) \cdot E^{ego}_{V}$, which helps evaluate model sensitivity to added parameters. ``CA, S2" represents standard cross-attention fusion, as shown in Eq.~\ref{eq:ca}.  ``CA + LearnResidual, S2" uses our MCFuse module. The ``S2" denotes that the entire network is trained only once (i.e., our second training stage). Results show: 1) Even simple methods like ``Add" and ``CA w/o Params" improve upon the base model, validating MCFuse’s core concept. 2) The notable drop with ``CA, S2" suggests caution when introducing new parameters into other pre-trained modules, motivating our two-stage training strategy. This strategy is further validated by comparing ``CA + LearnResidual, S2" with our MCFuse. 3) The improvement from ``CA, S2" to ``CA + LearnResidual, S2" supports protecting the more reliable visual prompt condition.
For the \textbf{residual weight} experiments, we compare our learnable approach with fixed weights $k_{lea}$ (0.2, 0.5, 0.8). Results indicate that learnable weight $k_{lea}$ reduces manual tuning while effectively adapting to appropriate values.
Regarding the \textbf{placement of MCFuse}, we compare put MCFuse ``Before XObjAlign" with ``After XObjAlign". Results show the latter is more effective, likely because applying the alignment loss $\mathcal{L}_{XObj}$ upon MCFuse may skew optimization toward better fusion objectives.

\subsection{More Ablations on XObjAlign}
Additionally, we provide further ablation studies on the XObjAlign module, including various metrics for computing the consistency constraint and different weights for $\mathcal{L}_{XObj}$.
Particularly, as in Tab.~\ref{tab:supp-tab3}, we first compare the performance of using the Euclidean loss for XObjAlign with that of using cosine similarity, and then compare different weights for composing the sublosses $\mathcal{L}_{mask}$ and $\mathcal{L}_{XObj}$. Note that only the XObjAlign module is applied.

\begin{table}[h]
    \centering
\resizebox{0.45\textwidth}{!}{
\begin{tabular}{l|l|l}
\hline
\textbf{Ablation Factor}      & \textbf{Method} & \textbf{IoU}$\uparrow$\\ \hline
-  & Base PSALM~\cite{zhang2024psalm} & 39.7 \\ \hline
\textbf{Metrics for XObjAlign} &  Cosine      &     42.5         \\ \cline{2-3}
                              & \cellcolor{cyan!10}Euclidean (Ours)         &   \cellcolor{cyan!10}\textbf{43.8}          \\ \hline
\textbf{Loss Weights} &  $\mathcal{L}$ =  $\mathcal{L}_{mask} + 0.2* \mathcal{L}_{XObj}$  & 41.2  \\  \cline{2-3}
                             &    $\mathcal{L}$ =  $\mathcal{L}_{mask} + 0.5* \mathcal{L}_{XObj}$    &   42.0           \\      \cline{2-3}
                            & $\mathcal{L}$ =  $\cellcolor{cyan!10} \mathcal{L}_{mask} + \mathcal{L}_{XObj}$  (Ours)          &\cellcolor{cyan!10}\textbf{43.8}            \\
                              \cline{2-3}
                              &    $\mathcal{L}$ =  $\mathcal{L}_{mask} + 10*\mathcal{L}_{XObj}$              &   40.3           \\
                               \bottomrule
\end{tabular}}
\caption{More ablation studies on XObjAlign and loss functions. Models are trained on Ego2Exo Small TrainSet.}
\label{tab:supp-tab3}
\end{table}

Results show that: 1) Compared to the cosine similarity, the Euclidean loss is better. However, both of them clearly outperform the base PSALM. 2) Among different choices, the default one $\mathcal{L} = \mathcal{L}_{mask} + \mathcal{L}_{XObj}$ achieves the best result.
The consistent improvement over the baseline observed across other weight ratios further demonstrates the robust positive impact of our XObjAlign module.

\subsection{First-Frame Query Evaluation}
Ego-Exo4D provides frame-level correspondence between ego and exo videos, while such extensive annotation might be impractical in the real world. Thus, we also test ObjectRelator under a more realistic scenario where only a first-frame query is provided. To adapt our model to this setting, we introduce a memory mechanism during inference. Taking Ego2Exo as an example, beyond the first frame—where the ego object mask serves as the query—the predicted results from previous exo frames are used as queries for subsequent exo frames. The results are presented in Tab.~\ref{tab:abla-track}.

\begin{figure*}[t!]
    \centering
       {\includegraphics[width=.99\linewidth]{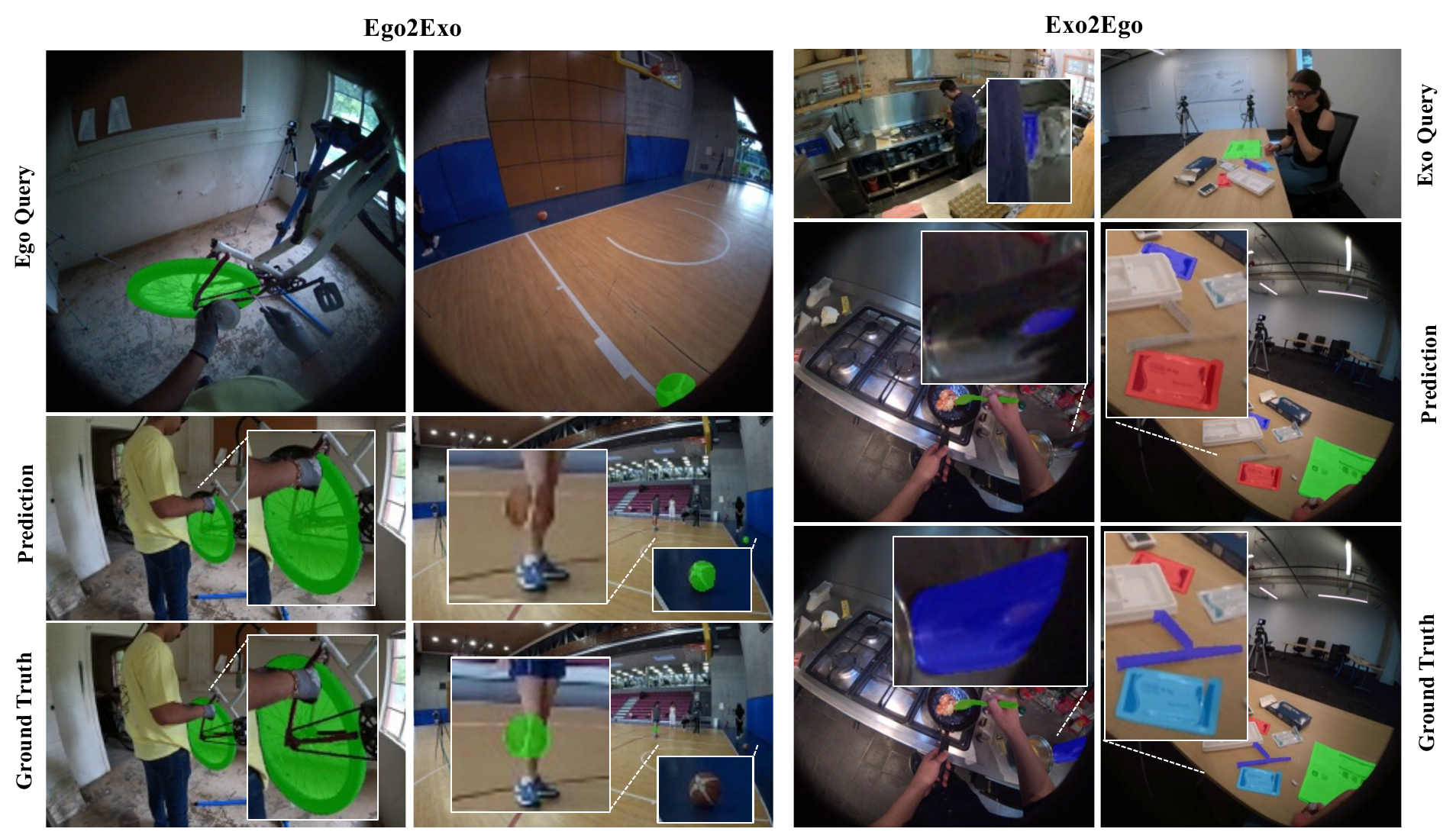}}
    \caption{Visualization of failure cases. 
    }
    \label{fig:vis-supp-fail}
\end{figure*}

\begin{table}[h]
    \centering 
    \setlength\tabcolsep{1.5pt}
    \setlength{\extrarowheight}{0.5pt}
    \scalebox{0.85}{
    \begin{tabular}{l|ccc}
    \toprule[0.95pt]
    \textbf{Method} & \textbf{Testing Query} &  \textbf{Ego2Exo}$\uparrow$& \textbf{Exo2Ego}$\uparrow$ \\ \midrule[0.6pt]
    \textbf{ObjectRelator} & \cellcolor{cyan!10} Frame Level (Ours) & \cellcolor{cyan!10}44.3 & \cellcolor{cyan!10}49.2 \\  \hline
    & 1-st Frame + Memory  & 37.0  & 43.7  \\ 
    \bottomrule[0.95pt]
    \end{tabular}
    }
    \caption{ObjectRelator with all frame-level queries vs. only 1st frame query.  Model trained on Small TrainSet is tested. }
    \label{tab:abla-track}
\end{table}

Results show that though not as good as the frame-level queries, our ObjectRelator could still work under this case.

\subsection{More Visualization Results}
\begin{figure*}[h]
    \centering
       {\includegraphics[width=1.\linewidth]{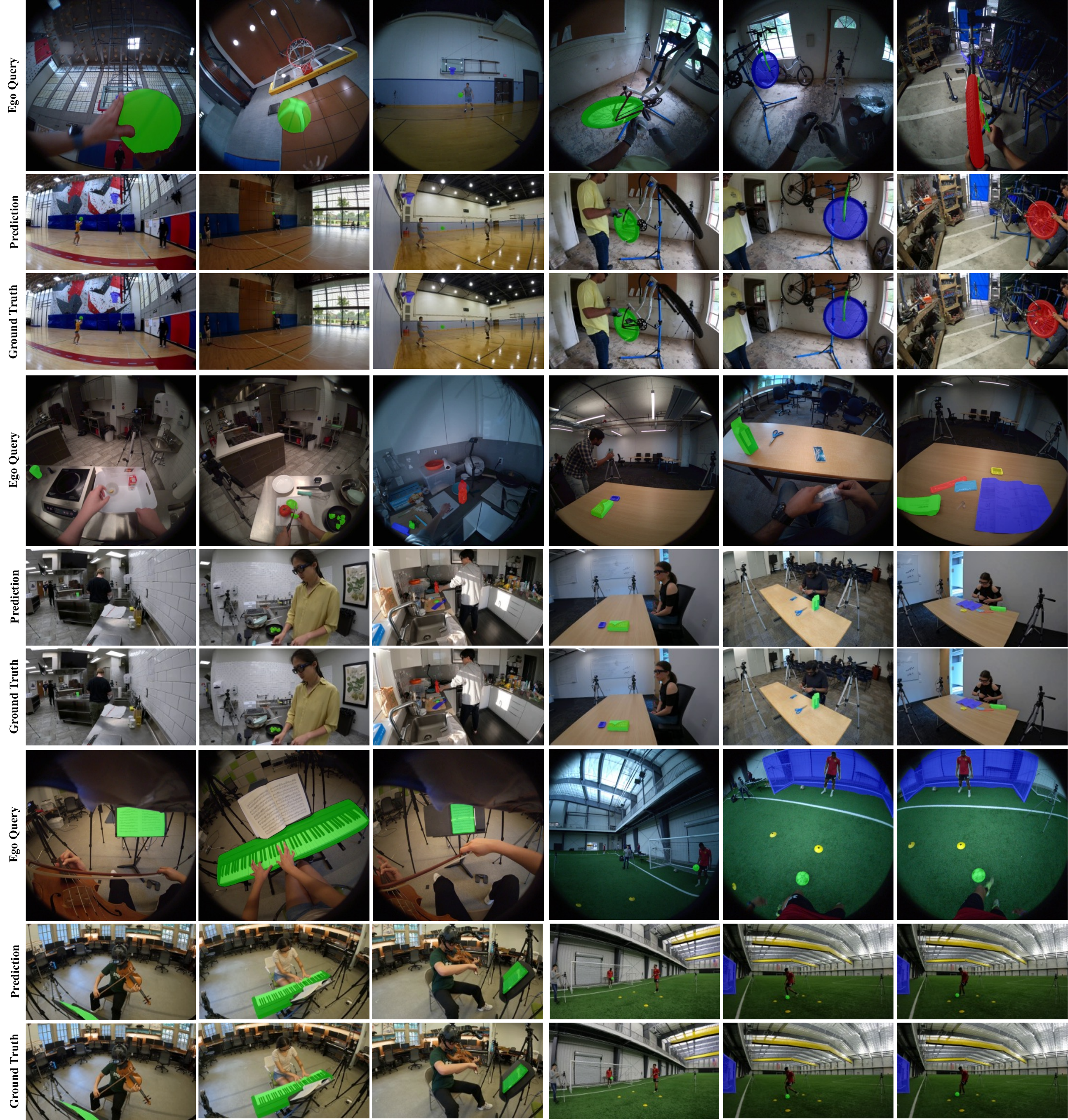}}
    \caption{Ego2Exo visualization results (ego query, predictions, and ground truth). Ego2Exo model trained on SmallTrain set is used.}
    \label{fig:vis-supp2}
\end{figure*}

\begin{figure*}[h]
    \centering
       {\includegraphics[width=1.\linewidth]{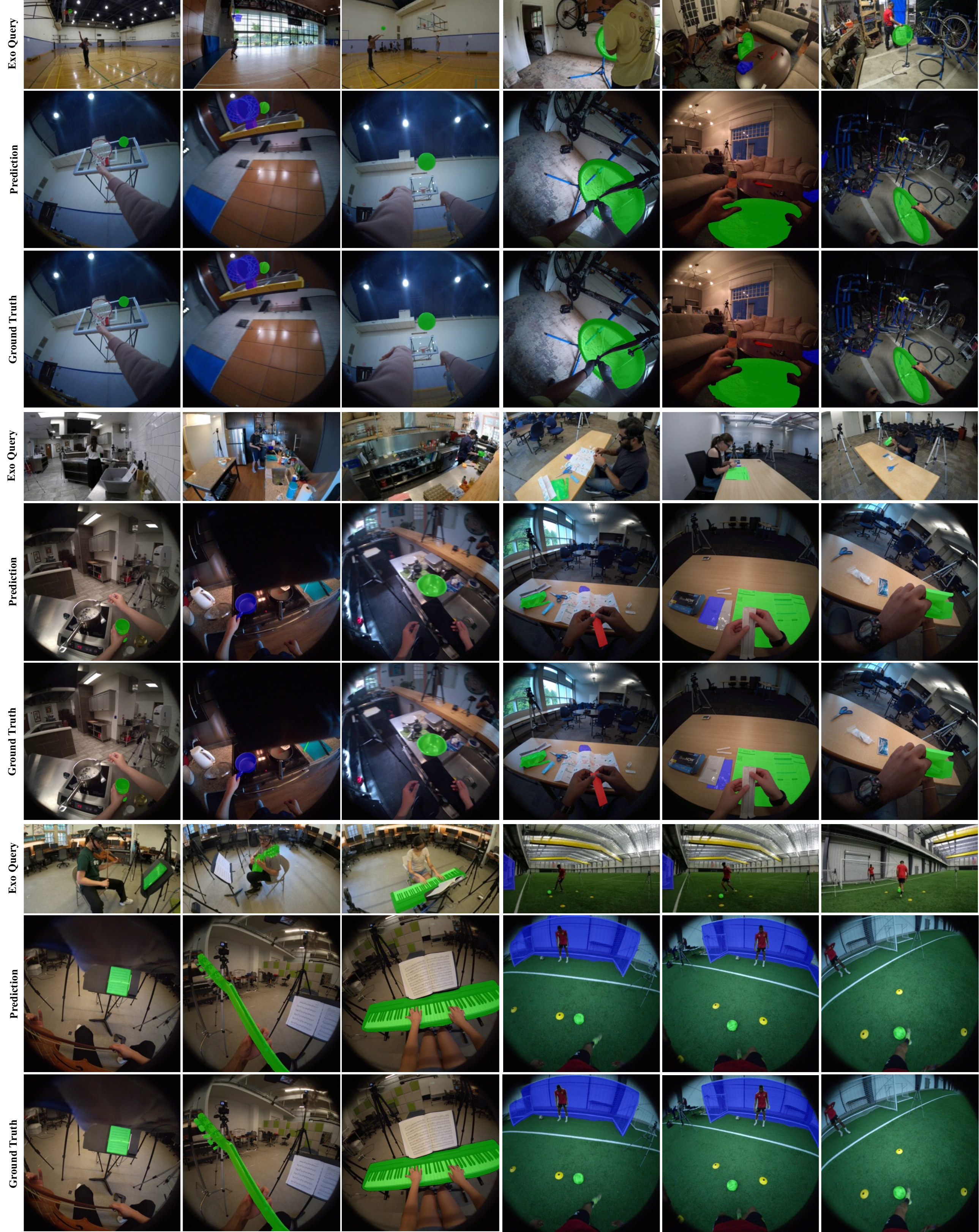}}
    \caption{Exo2Ego visualization results (exo query, predictions, and ground truth). Exo2Ego model trained on SmallTrain set is used.}
    \label{fig:vis-supp3}
\end{figure*}

\begin{figure*}[h]
    \centering
         \vspace{-0.05in}
       {\includegraphics[width=1.\linewidth]{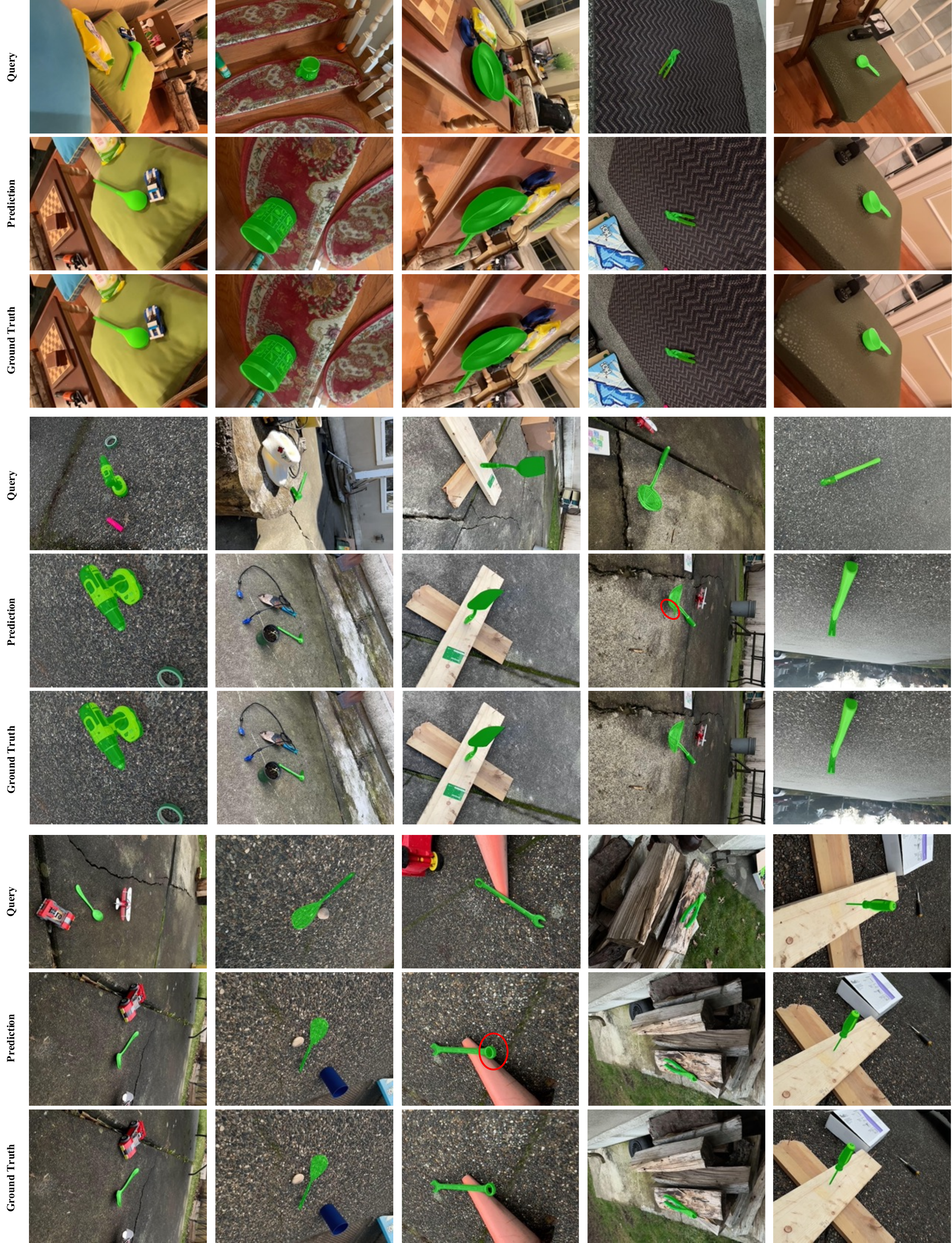}}
       \vspace{-0.25in}
    \caption{HANDAL-X visualization results (query, predictions, and ground truth). }
    \label{fig:vis-handal}
\end{figure*}

Due to space limitations, only a few examples are presented in Fig.\ref{fig:vis-overall}.
We include additional visualization results in Fig.\ref{fig:vis-supp2} (Ego2Exo) and Fig.~\ref{fig:vis-supp3} (Exo2Ego). For each subtask, we showcase diverse examples spanning all six scenarios: Cooking, BikeRepair, Health, Music, Basketball, and Soccer.
Results demonstrate that our proposed ObjectRelator effectively segments cross-view objects in most cases, producing results that closely align with ground truth annotations. This highlights the robustness of our method in handling diverse scenarios, various object categories, and challenging cases, such as occlusion and novel viewpoints. We also provide visualization results (Fig.~\ref{fig:vis-handal}) on HANDAL-X. 15 examples across diverse categories are demonstrated. Our model accurately predicts masks matching the ground truth, with only a few failures highlighted in red circles.

\section{Failure Cases}\label{sec:supp-fail}
We also analyze the failure cases produced by our method, with typical examples summarized in Fig.~\ref{fig:vis-supp-fail}. Results indicate that our method struggles in several scenarios: it fails to generate a complete mask when the object's surface is discontinuous or blends closely with the background (first and third columns), incorrectly identifies the object when multiple similar objects are present in the scene (second and fourth columns), and occasionally misses some objects entirely (fourth column).

\section{Limitations and Future Work}\label{sec:supp-diss}
In this paper, we propose ObjectRelator, a method designed to understand cross-view object relationships in terms of segmentation, validated on ego-exo perspectives and a relatively easier cross-view dataset.  
The approach primarily consists of multimodal condition fusion and SSL-based cross-view object alignment, built on top of a frame-level multimodal segmentation model. While our method achieves SOTA results, as demonstrated in Fig.~\ref{fig:vis-supp-fail}, there still remains significant room for improvement. This underscores the substantial challenges and rich opportunities in addressing ego-exo object correspondence. Looking ahead, we plan to explore the integration of temporal information to better capture object dynamics.

